\documentclass[conference]{IEEEtran}
\IEEEoverridecommandlockouts
\usepackage{cite}
\usepackage{amsmath,amssymb,amsfonts}
\usepackage{algorithmic}
\usepackage{graphicx}
\usepackage{textcomp}
\usepackage{url}
\usepackage{breakurl}
\usepackage{xcolor}
\def\BibTeX{{\rm B\kern-.05em{\sc i\kern-.025em b}\kern-.08em
    T\kern-.1667em\lower.7ex\hbox{E}\kern-.125emX}}
\begin{document}

\title{Early Detection of Coffee Leaf Rust Through Convolutional Neural Networks Trained on Low-Resolution Images\\}

\author{
\IEEEauthorblockN{Angelly Cabrera}
\IEEEauthorblockA{University of Southern California}
\IEEEauthorblockA{Signal Analysis and Interpretation Lab}
\and
\IEEEauthorblockN{Kleanthis Avramidis} 
\IEEEauthorblockA{University of Southern California}
\IEEEauthorblockA{Signal Analysis and Interpretation Lab}
\and
\IEEEauthorblockN{Shrikanth Narayanan} 
\IEEEauthorblockA{University of Southern California}
\IEEEauthorblockA{Signal Analysis and Interpretation Lab}
}

\maketitle

\begin{abstract}
Coffee leaf rust, a foliar disease caused by the fungus Hemileia vastatrix, poses a major threat to coffee production, especially in Central America. Climate change further aggravates this issue, as it shortens the latency period between initial infection and the emergence of visible symptoms in diseases like leaf rust. Shortened latency periods can lead to more severe plant epidemics and faster spread of diseases. There is, hence, an urgent need for effective disease management strategies. To address these challenges, we explore the potential of deep learning models for enhancing early disease detection. However, deep learning models require extensive processing power and large amounts of data for model training, resources that are typically scarce. To overcome these barriers, we propose a preprocessing technique that involves convolving training images with a high-pass filter to enhance lesion-leaf contrast, significantly improving model efficacy in resource-limited environments. This method and our model demonstrated a strong performance, achieving over 90\% across all evaluation metrics—including precision, recall, F1-score, and the Dice coefficient. Our experiments show that this approach outperforms other methods, including two different image preprocessing techniques and using unaltered, full-color images. 
\end{abstract}

\section{Introduction}
Central America's volcanic topography and tropical climate make this isthmus between North and South America an ideal environment for coffee cultivation and, conversely, the spread of leaf rust. Coffee leaf rust, affecting more than 70 countries, has been particularly devastating in Latin America, resulting in economic losses of up to \$1 billion and the displacement of 250,000 agricultural jobs, hence becoming a growing motivator of mass migrations [1]. Despite its relatively small size, Guatemala is renowned for its diverse micro-climates across eight distinct regions, each offering varying altitudes and unique growing conditions. These micro-climates contribute to cultivating various coffee types, such as Catuai, which, while having a high yield potential, is highly susceptible to leaf rust. This susceptibility can significantly reduce yields and increase plant mortality [2]. As climate change continues to impact this region, rising temperatures will lead to shorter latency periods for diseases like coffee leaf rust, exacerbating the threat to coffee production—a critical component of Central America's economy, accounting for over 20\% of its exports [3].

Research conducted by the Center for Tropical Agriculture (CIAT) and the International Coffee Organization (ICO) emphasizes the urgent need to adapt technological solutions to aid farmers effectively at the field scale [2]. Early disease detection is critical as it provides farmers the necessary time to counteract the rapidly shortening latency periods, thereby preventing widespread outbreaks. However, a significant gap in current research is the insufficient focus on early-stage leaf rust, which manifests as white dots less than 1mm in size, as shown in Figure 1 [4]. To address this gap, we propose training a Convolutional Neural Network (CNN) to target early-stage leaf rust by applying image convolution to the Coffee Leaf Disease dataset [5], which consists of early-stage leaf rust on Arabica variety coffee leaves. This method aims to enhance the accuracy and timeliness of leaf rust detection, potentially improving proactive disease management practices in coffee cultivation.

\begin{figure}[htbp]
\centering
\includegraphics[width=\linewidth]{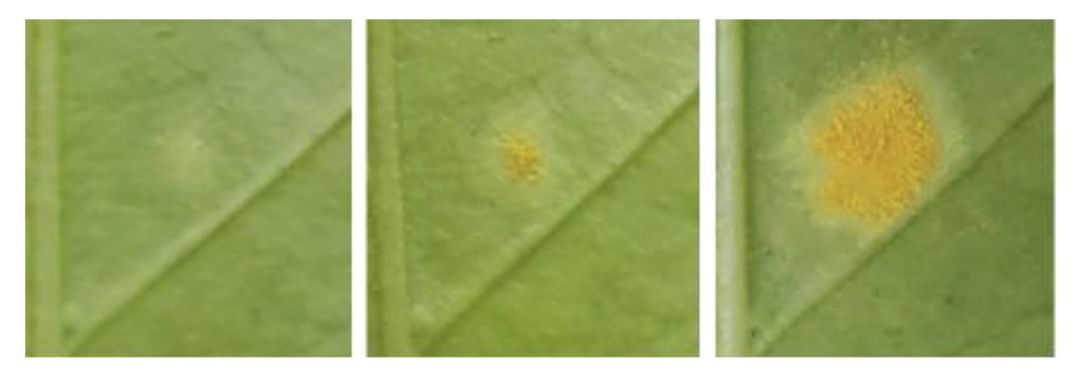}
\caption{Stages of leaf rust beginning from the lesion emergence (Source: [4]).}
\label{fig}
\end{figure}

\subsection{Analysis of Existing Solutions}
Current methods for leaf disease detection include remote sensing, wireless sensor networks, and deep learning [6]. Remote sensing techniques primarily use satellite multispectral images; however, the resolution of these images often fails to detect early-stage diseases in individual plants. Alternatively, machine learning offers a less resource-intensive, non-invasive approach through image classification. Within deep learning, most existing research on leaf rust detection uses high-resolution RGB or HSV images for training [7]. These approaches yield varied results and involve different levels of model complexity. Although CNNs are widely used for plant disease detection and phenotyping, they face significant challenges due to the scarcity of high-quality data, particularly in the early stages of leaf rust [8]. To address these challenges, we propose an alternative approach, through which the images are first convolved with a high-pass kernel to aid feature extraction, followed by image downsizing. This method aims to improve the detection capabilities of CNNs under data scarcity and low image quality conditions, thus offering a more effective solution for early detection of leaf rust, particularly in resource-constrained regions.

A similar approach has been made to training deep learning models on low-resolution images, such as the DiffuCNN model for tobacco plants [9]. This model integrates a diffusion-based resolution enhancement module to improve image clarity and detail, making it more effective for detecting and grading tobacco diseases in low-resolution, complex agricultural scenes. The DiffuCNN model significantly enhances computational efficiency and is adaptable for deployment across various devices with limited processing power. Additionally, research on low-resolution images for plant disease detection, such as the work by [10], demonstrates that low-resolution images can effectively retain essential information while reducing computational load. This approach not only makes disease detection more accessible and affordable for farmers but also enables real-time processing on devices with limited resources.

\section{PROBLEM FORMULATION}

\subsection{Dataset}
Images acquired from the Coffee Leaf Disease dataset [5], later aggregated with images from the Disease and Pest in Coffee Leaves dataset [11], were convolved with a high-pass kernel to output a new set of images with a sharp contrast between leaf edges and rust. These datasets consist of coffee leaves from India and Brazil, which were manually collected from farms and already exhibiting various diseases. For this study, we specifically focused on extracting images that show leaf rust. Figure 2 shows a sample image before and after applying the suggested augmentation, where lighter pixel values are inverted to contrast surrounding shadows. This kernel was designed to produce strong negative responses for vertical edges transitioning from low to high pixel values. The images were then converted to grayscale and downscaled to reduce computational complexity. Initially, the images were resized to 64x64 pixels to mimic the resolution of well-known datasets such as MNIST and Fashion MNIST and reduce the computational load as much as possible. Observing the stagnating performance of the CNN, the images were later re-scaled to 128x128, which achieved significantly better performance. Reducing the image resolution balances information retention with computation efficiency, making the model more adaptable for deployment across various devices, including those with limited processing power. Figure 3 illustrates the appearance of healthy and infected leaves after these processing steps.

\begin{figure}[htbp]
\centering
\includegraphics[width=\linewidth]{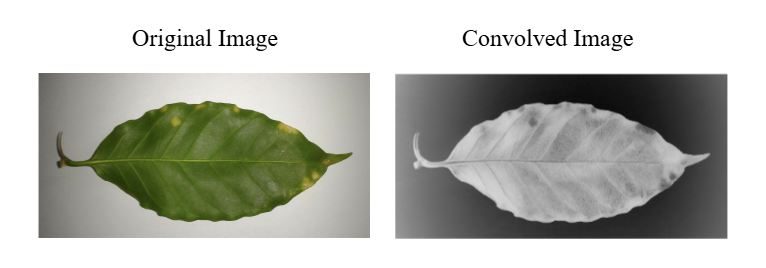}
\caption{Image pre-convolution (left) and post-convolution (right).}
\label{fig}
\end{figure}

\begin{figure}[htbp]
\centering
\includegraphics[width=\linewidth]{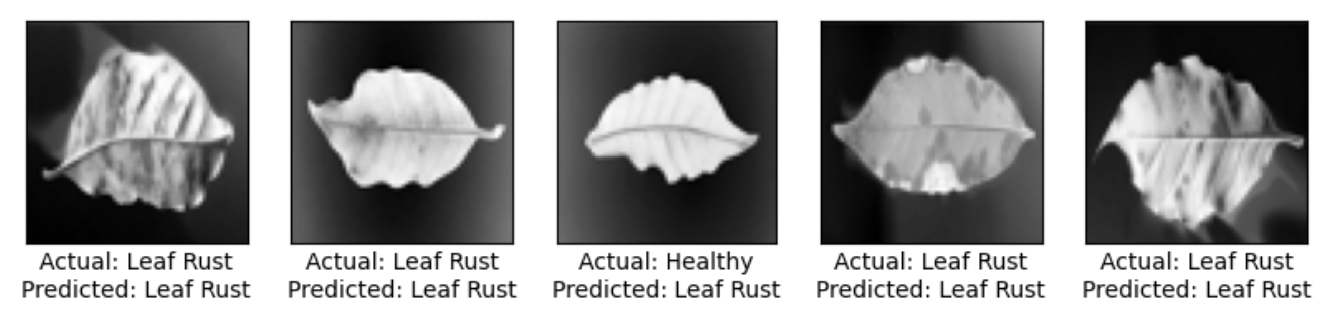}
\caption{Test images post-convolution and downscaling with model predictions.}
\label{fig}
\end{figure}

\subsection{Image Convolution}
As discussed previously, existing deep-learning studies typically train CNNs on high-resolution RGB images. However, detecting early-stage leaf rust is challenging due to the color similarity between the lesion and the leaf. To address this, we experimented with several image processing techniques and found that edge detection through high-pass filtering rendered the best lesion-leaf contrast. The images were first converted to grayscale and then convolved with the kernel, as illustrated in Figure 4. After applying convolution, the images were normalized to standardize pixel values ranging from 0 to 255. We also experimented with histogram equalization on the convolved and grayscale images, demonstrated in Figure 5. Histogram equalization adjusts the image's contrast by modifying its intensity distribution to achieve a desired histogram shape. However, equalization did not improve the contrast between the leaf textures and rust for grayscale images. Additionally, equalizing the convolved images created image blurs that could confuse the model with unnecessary details. 

\begin{figure}[htbp]
\centering
\includegraphics[width=\linewidth]{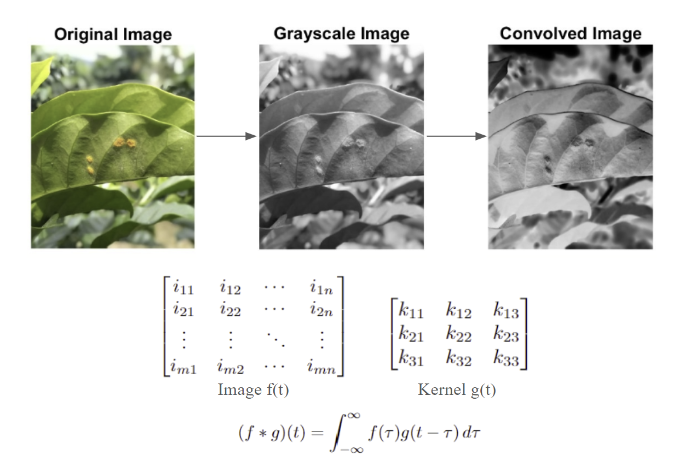}
\caption{Convolution process on a sample image.}
\label{fig}
\end{figure}

\begin{figure}[htbp]
\centering
\includegraphics[width=\linewidth]{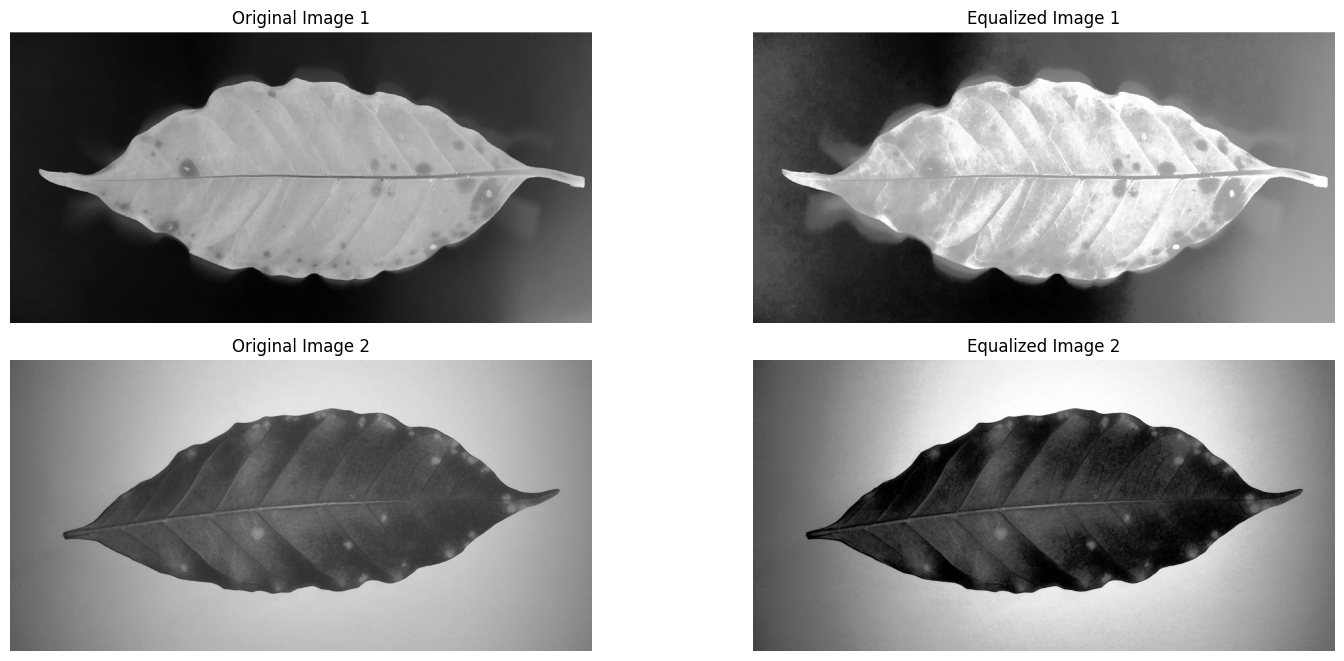}
\caption{Histogram equalization on convolved (top) and grayscale images.}
\label{fig}
\end{figure}

\subsection{Model}
Our proposed CNN comprises nine layers, beginning with two convolution layers, then a max pooling, batch normalization, flattening layer, and finishing with four dense layers to integrate features from preceding layers and make final class predictions. Incorporating batch normalization significantly enhances model training by standardizing the outputs of previous layers, reducing the number of training epochs required, and decreasing the likelihood of the model learning noise from the training data—a common source of overfitting. The model effectively retains its learning capacity without sacrificing performance by substituting batch normalization for dropout layers. In addition to normalization, we used early stopping if there was no improvement for 60 consecutive iterations, with a cap of 200 epochs on training. These regularization techniques, paired with dataset resizing, resulted in a 5\% increase in precision and recall metrics.

\begin{figure}[htbp]
\centering
\includegraphics[width=\linewidth]{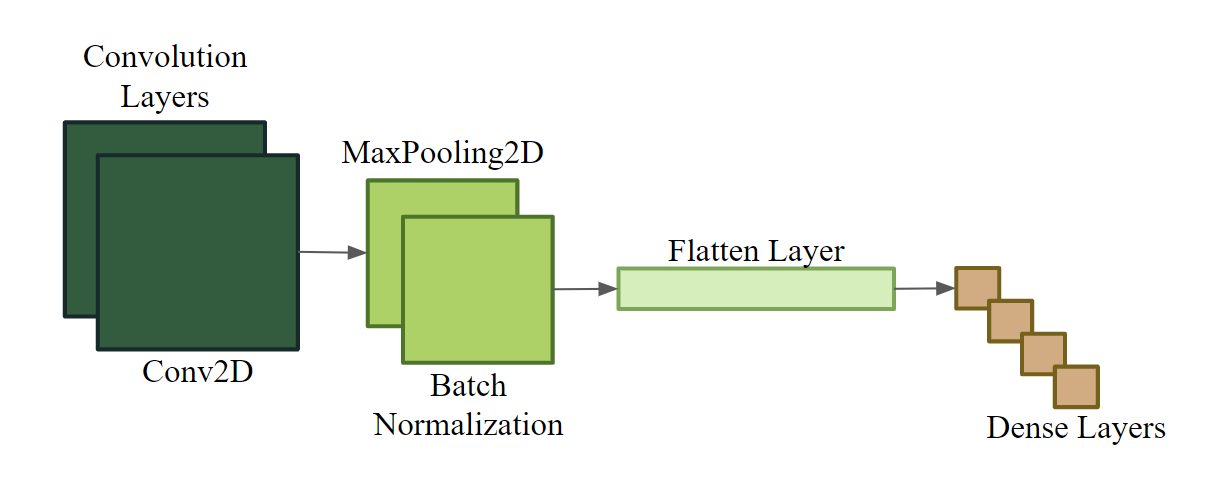}
\caption{Architecture of Convolutional Neural Network.}
\label{fig}
\end{figure}

\section{RESULTS}
The CNN's performance was assessed using precision, recall, F1-score, and the Dice coefficient. Precision and recall are particularly crucial for this application. High precision minimizes missed diagnoses, preventing delayed treatments and crop losses. Conversely, high recall reduces the likelihood of healthy leaves being misclassified as infected, which helps avoid unnecessary interventions and optimizes resource use. The F1-score provides a balanced measure of precision and recall, while the Dice coefficient assesses the similarity between predicted and actual classes. As observed in Table 1, the CNN demonstrated a robust performance across all tested resolutions, demonstrating its effectiveness on low-resolution images. The highest-resolution images achieved the best performance across all metrics, confirming that higher resolutions provide sufficient detail to refine model predictions. Notably, the 128x128 resolution provides an optimal balance, offering enough detail for accurate detection while requiring less computational power for processing.

\begin{table}[htbp]
\caption{Performance Metrics for Different Image Resolutions}
\begin{center}
\begin{tabular}{|c|c|c|c|c|}
\hline
\textbf{Image Resolution} & \textbf{\textit{Precision}} & \textbf{\textit{Recall}} & \textbf{\textit{F1}} & \textbf{\textit{Dice Coefficient}} \\
\hline
64x64 & 0.857 & 0.856 & 0.856 & 0.869 \\
\hline
84x84 & 0.880 & 0.878 & 0.879 & 0.883 \\
\hline
128x128 & 0.939 & 0.939 & 0.939 & 0.942 \\
\hline
\end{tabular}
\label{tab1}
\end{center}
\end{table}

The CNN's performance was also evaluated using four different preprocessing methods, including two that did not utilize edge detection, to compare the effectiveness of convolution on the dataset before training the model. As shown in Table 2, methods incorporating edge detection achieved higher precision and Dice coefficient values, indicating a greater similarity between the model's predictions and the actual classes. Specifically, the method combining edge detection with histogram equalization recorded the second-highest performance. Nevertheless, the difference between this approach and the highest-performing method is significant. In contrast, the absence of edge detection in grayscale images resulted in the lowest precision and Dice coefficient values. 

\begin{table}[htbp]
\caption{Performance Metrics for Different Preprocessing Methods}
\begin{center}
\begin{tabular}{|c|c|c|c|c|}
\hline
\textbf{Preprocessing Method} & \textbf{\textit{Precision}} & \textbf{\textit{Recall}} & \textbf{\textit{F1}} & \textbf{\textit{Dice}} \\
\hline
Edge Detection (Grayscale) & 0.939 & 0.939 & 0.939 & 0.942 \\
\hline
No Edge Detection (Grayscale) & 0.117 & 0.341 & 0.174 & 0.000 \\
\hline
No Edge Detection (Color) & 0.627 & 0.745 & 0.681 & 0.681 \\
\hline
Edge Detection Equalization & 0.784 & 0.785 & 0.785 & 0.796 \\
\hline
\end{tabular}
\label{tab2}
\end{center}
\end{table}

\section{CONCLUSION}
\subsection{Discussion of Results}
The results demonstrate that the CNN achieved performances above 90\% across all evaluation metrics, proving its potential to learn from low-resolution images and adapt to the realities of environments that lack sophisticated, high-resolution imaging equipment. When reduced to low-resolution, images of early leaf rust, which often appear as tiny white specks, become indistinguishable from other features of the leaf in RGB images. By applying a high-pass filter during preprocessing, we enhance the contrast between rust lesions and other aspects of the leaf such as edges. Not only does this approach highlight the initial symptoms in lower-quality images, but it also provides a 'shortcut' for the model. This technique allows the CNN to focus on significant features from the outset, rather than needing to learn to recognize these subtle details from a larger volume of data. This preprocessing acts as a knowledge-based enhancement, streamlining the model's training process and improving its ability to recognize the early indicators of the disease efficiently.

Despite the clear advantages of this approach, it presents several challenges that need to be considered. Initially, the intention was to employ a pre-trained model like AlexNet [12]. However, this strategy was impractical due to the relatively small dataset size, comprising just over 600 images, which is substantially fewer than those used in comparable research. Additionally, training the model on high-resolution images would compromise its performance in accurately identifying the early stages of leaf rust, hence emphasizing the need for tailored approaches in resource-limited settings.

\subsection{Future Work}
While existing deep learning methods are becoming increasingly effective, their dependency on large amounts of well-labeled data presents significant challenges due to data collection's labor-intensive and time-consuming nature. To mitigate this, future research could investigate applying unsupervised labeling techniques, such as k-nearest neighbors (KNNs), for scoring the severity of leaf rust [13]. This approach could reduce reliance on extensive labeled datasets. Furthermore, Generative Adversarial Networks (GANs) provide a promising avenue by generating new, diverse training data, which could enhance the robustness of detection methods. Such advancements are crucial for adapting to the rapidly shortening latency periods and the continuous evolution of new leaf rust strains, offering a more adaptive and efficient solution to this agricultural challenge.

\vspace{12pt}


\begin{thebibliography}{00}
\bibitem{b1} A. Koutouleas, D. B. Collinge, and E. Boa, “The coffee leaf rust pandemic: An ever‐present danger to coffee production,” Plant pathology, vol. 73, no. 3, pp. 522–534, 2024, doi: 10.1111/ppa.13846.
\bibitem{b2} J. Avelino et al., “The coffee rust crises in Colombia and Central America (2008–2013): impacts, plausible causes and proposed solutions,” Food security, vol. 7, no. 2, pp. 303–321, 2015, doi: 10.1007/s12571-015-0446-9.
\bibitem{b3} A. P. Marcos, N. L. Silva Rodovalho and A. R. Backes, ”Coffee Leaf Rust Detection Using Convolutional Neural Network,” 2019 XV Workshop de Visao Computacional (WVC), Sao Bernardo do Campo, Brazil, 2019, pp. 38-42, doi: 10.1109/WVC.2019.8876931.
\bibitem{b4} I. Merle, P. Tixier, E. de M. Virginio Filho, C. Cilas, and J. Avelino, “Forecast models of coffee leaf rust symptoms and signs based on identified microclimatic combinations in coffee-based agroforestry systems in Costa Rica,” Crop protection, vol. 130, pp. 105046-, 2020, doi: 10.1016/j.cropro.2019.105046.
\bibitem{b5} G. Dutta, ”Coffee Leaf Diseases,” Kaggle, Dataset, May 2023. [Online]. Available: \url{https://www.kaggle.com/datasets/gauravduttakiit/coffee-leaf-diseases}.
\bibitem{b6} D. Velásquez, A. Sánchez, S. Sarmiento, M. Toro, M. Maiza, and B. Sierra, “A method for detecting coffee leaf rust through wireless sensor networks, remote sensing, and deep learning: Case study of the Caturra variety in Colombia,” Applied sciences, vol. 10, no. 2, pp. 697-, 2020, doi: 10.3390/app10020697.
\bibitem{b7} F. G. Waldamichael, T. G. Debelee, and Y. M. Ayano, “Coffee disease detection using a robust HSV color‐based segmentation and transfer learning for use on smartphones,” International journal of intelligent systems, vol. 37, no. 8, pp. 4967–4993, 2022, doi: 10.1002/int.22747.
\bibitem{b8} D. Novtahaning, H. A. Shah, and J. M. Kang, “Deep Learning Ensemble-Based Automated and High-Performing Recognition of Coffee Leaf Disease,” Agriculture (Basel), vol. 12, no. 11, pp. 1909-, 2022, doi: 10.3390/agriculture12111909.
\bibitem{b9} H. Xiong et al., “DiffuCNN: Tobacco Disease Identification and Grading Model in Low-Resolution Complex Agricultural Scenes,” Agriculture (Basel), vol. 14, no. 2, pp. 318-, 2024, doi: 10.3390/agriculture14020318.
\bibitem{b10} J. I. Chen, H. Wang, K.-L. Du, and V. Suma, “Deep-CNN for Plant Disease Diagnosis Using Low Resolution Leaf Images,” in Machine Learning and Autonomous Systems, vol. 269, Singapore: Springer Singapore Pte. Limited, 2022, pp. 459–469. doi: \url{10.1007/978-981-16-7996-4_33}.
\bibitem{b11} Alvaro Leandro Cavalcante Carneiro, L. de Brito Silva, and Marisa Silveira Almeida Renaud Faulin, “Artificial intelligence for detection and quantification of rust and leaf miner in coffee crop,” arXiv.org, 2021, doi: \url{10.48550/arxiv.2103.11241}.
\bibitem{b12} A. Krizhevsky, I. Sutskever, and G. Hinton, ImageNet classification with deep convolutional neural networks, vol. 60, no. 6. NEW YORK: ACM, 2017, pp. 84–90. doi: \url{10.1145/3065386}.
\bibitem{b13} T. C. Pham, V. D. Nguyen, C. H. Le, M. Packianather, and V. D. Hoang, “Artificial intelligence-based solutions for coffee leaf disease classification,” IOP Conference Series: Earth and Environmental Science, vol. 1278, no. 1, pp. 12004-, 2023, doi: \url{10.1088/1755-1315/1278/1/012004}.
\end{thebibliography}
\end{document}